\pdfoutput=1
\documentclass[11pt]{article}
\usepackage[table]{xcolor}
\usepackage{acl}
\usepackage{multirow, makecell}
\usepackage{tabularx}
\usepackage{times}
\usepackage{latexsym}
\usepackage{booktabs,siunitx,multirow}
\usepackage{amsmath}
\usepackage[T1]{fontenc}
\usepackage{graphicx} 
\usepackage{subcaption}
\usepackage{booktabs}
\usepackage{todonotes}
\usepackage{nicematrix,tikz}
\definecolor{Gray}{gray}{0.85}
\definecolor{LightCyan}{rgb}{0.88,1,1}
\usepackage[table]{xcolor}
\newcolumntype{a}{>{\columncolor{Gray}}c}
\newcolumntype{b}{>{\columncolor{white}}c}
\usepackage{pgfplots} 
\usetikzlibrary{patterns}
\makeatletter
\def\thanks#1{\protected@xdef\@thanks{\@thanks
        \protect\footnotetext{#1}}}
\makeatother
\usepackage[utf8]{inputenc}
\usepackage{xspace}
\usepackage{microtype}
\usepackage[usestackEOL]{stackengine}
\newcolumntype{P}[1]{>{\centering\arraybackslash}p{#1}}

\title{Fake News Detectors are Biased against \\Texts Generated by Large Language Models}

\author{
    Jinyan Su$^{1\dagger}$,
    Terry Yue Zhuo$^{2\dagger}$,
    Jonibek Mansurov$^1$,
    Di Wang$^3$,
    Preslav Nakov$^1$\\
    $^1$Mohamed bin Zayed University of Artificial Intelligence\\
    $^2$Monash University and CSIRO's Data61\\
    $^3$King Abdullah University of Science and Technology \\
    \texttt{\{Jinyan.Su, preslav.nakov\}@mbzuai.ac.ae}\\
    \texttt{terry.zhuo@monash.edu, di.wang@kaust.edu.sa}\\
    \thanks{$\dagger$: Equal contribution.}
}

\begin{document}
\maketitle
\begin{abstract}
The spread of fake news has emerged as a critical challenge, undermining trust and posing threats to society. In the era of Large Language Models (LLMs), the capability to generate believable fake content has intensified these concerns. In this study, we present a novel paradigm to evaluate fake news detectors in scenarios involving both human-written and LLM-generated misinformation. Intriguingly, our findings reveal a significant bias in many existing detectors: they are more prone to flagging LLM-generated content as fake news while often misclassifying human-written fake news as genuine. This unexpected bias appears to arise from distinct linguistic patterns inherent to LLM outputs. To address this, we introduce a mitigation strategy that leverages adversarial training with LLM-paraphrased genuine news. The resulting model yielded marked improvements in detection accuracy for both human and LLM-generated news. To further catalyze research in this domain, we release two comprehensive datasets, \texttt{GossipCop++} and \texttt{PolitiFact++}, thus amalgamating human-validated articles with LLM-generated fake and real news.

\end{abstract}

\section{Introduction}
% { \color{red}{To be added: Human evaluation}}
% { \color{red}{Real news rephrased by machine}}
% Main structure in Introduction:

% LLM improves productivity--> both main stream news web and fake news use LLMs to help them write articles --> we should not only consider human fake news, but also machine generated one. (How would the fake news detector perform under the this condition?
% In this paper, we study the fake news detection problem under the setting that the fake news are not only generated by human, but also by LLMs such as ChatGPT. Surprisingly, we found that fake news detectors are more likely to correctly identify machine generated fake news, which implies that it might be biased towards machine generated news articles and using whether an article is generated by machine as a main decision rule when classifying fake news. 

%Fake news -> human written -> costly -> LLMs make it worse -> detect such LLM-generated text -> main RQ -> findings on bias -> why -> mitigate
\begin{quote}
\textit{In an age of universal deceit, telling the truth is a revolutionary act.}
\begin{flushright}
--- George Orwell
\end{flushright}
\end{quote}
The dissemination of false information can cause
chaos, hatred, and trust issues, and can eventually
hinder the development of society as a whole~\cite{wasserman2019exploratory}. Among them, fake news is often used to manipulate certain populations and had a catastrophic impact on multiple events, such as Brexit~\cite{bastos2019brexit},
the COVID-19 pandemic~\cite{van2020inoculating}, and the 2022 Russian assault on Ukraine~\cite{mbah2022russian}. To spread such fake news, adversaries conventionally will deploy propaganda techniques and manually write the fake news~\cite{huang2022faking}.  

Creating convincing disinformation manually is a labor-intensive and time-consuming process, which may limit the scale and speed at which such content can be produced. This makes it less efficient and desirable for adversaries who aim for widespread and rapid dissemination of false information~\cite{zellers2019defending}. With the development of language models like GPT-2~\cite{radford2019language} and BART~\cite{lewis2019bart}, more and more adversaries tend to utilize these models to automate fake news curation, resulting in a surge in the amount of fake news~\cite{weidinger2021ethical}. The recent advances in large language models (LLMs) have exacerbated the situation, as their increased capabilities can generate more convincing and nuanced disinformation at an unprecedented scale~\cite{pan2023risk,zhuo2023exploring}. For instance, the emergence and application of LLMs~\cite{brown2020language, touvron2023llama, li2023starcoder} like GPT-3 and ChatGPT have markedly impacted the media landscape. From January 1, 2022, to April 1, 2023, there was a dramatic surge in synthetic articles, especially on misinformation news websites~\cite{hanley2023machine}. Relative to the previous year, there was an increase of 79.4\% in the production of synthetic news articles on mainstream websites. However, this pales compared to the astounding 342\% increase seen on misinformation-oriented sites over the same period.

With the increasing concerns that humans are likely deceived or misled by LLM-generated fake news, there is an urgent need to study how the era of LLMs can affect fake news detection. Previous works have only trained fake news detectors to detect human-written or language-model-generated fake news~\cite{figueira2017current,zellers2019defending,schuster2020limitations}. Compared to these studies, we consider a more realistic scenario where the detectors must identify both human-written and LLM-generated fake news. Intuitively, we add the same amount of LLM-generated fake news as human-written to the training and test sets. Different from \citet{zellers2019defending} and \citet{pagnoni2022threat} aiming to defend against synthetic fake news via specific designs, our goal is to examine the performance of generic fake news detectors in detecting naturally written fake news by LLMs and humans. To synthesize the natural fake news via LLMs, we design a systematic framework to instruct LLMs with identifiable structures. We choose ChatGPT as the backbone model, as it is one of the most representative instruction-tuned LLMs that can generate human-like context.

Throughout our experiments on various fake news detectors like BERT, RoBERTa and ELECTRA~\cite{khan2021benchmark}, we surprisingly find that they can detect LLM-generated fake news better than human-written ones, in contrast to previous concerns about the challenges of identifying LLM-generated fake news~\cite{pan2023risk}. To further understand this finding, we continue paraphrasing human-written real news via ChatGPT and evaluate whether the detectors can correctly identify both LLM-paraphrased and human-written real news. We find that fake news detectors perform much worse on LLM-paraphrased real news than human-written ones. Based on these observations, we conclude that \textbf{fake news detectors are biased towards LLM-generate texts} and tend to classify them as fake news regardless of their truthfulness.

To mitigate such biases, we first study whether fake news detectors may take `shortcuts' to learn the LLM-generated fake news. Inspired by content-based features of news articles\citet{horne2017just,norregaard2019nela}, we analyze the new landscape (NELA) features and provide several hypotheses based on the statistical evidence. We demonstrate that bias can be mitigated by training on selective features with two regression detectors. We further propose a debiasing technique for fake news detectors by leveraging adversarial training with LLM-paraphrased real news. We show that our approach can effectively mitigate the biases and narrow the performance gap between LLM-generated and human-written texts.

Our contributions can be summarized as follows:

% \item We evaluate fake news detectors performance under a more practical setting, where we have both human generated fake news and machine generated ones. Our experimentally show that fake news detectors are biased towards machine generated news articles and are more likely to classify machine generated articles as fake. 

% \item We provide extensive analysis and understanding about this observation through explainable feature analysis and provide mitigation to avoid such bias when doing fake news detections. 

% \item We complement \texttt{GossipCop++} with 4084 machine generated fake news and release the dataset for future research on fake news detection.

\begin{itemize}
\item We introduce a new and realistic setting for evaluating fake news detectors. In this scenario, detectors must identify both human-written and LLM-generated fake news. This reflects real-world situations more accurately, considering the increasing usage of LLMs in disseminating disinformation. Testing detectors against human and LLM-generated content allows us to assess their resilience and effectiveness in an evolving fake news landscape.

\item Our analysis uncovers surprising findings. Despite existing concerns about the ability of fake news detectors to identify LLM-generated fake news, we find these detectors demonstrate a bias. They disproportionately classify LLM-generated content as fake news, even when it is truthful.

\item We delve deeper into these observations, suggesting potential explanations for the detected bias via content-based NELA features. We propose that these detectors may learn `shortcuts', identifying fake news based on unique linguistic features in LLM-generated texts.

\item On the basis of this bias, we develop a mitigation technique leveraging adversarial training~\cite{bai2021recent} with LLM-paraphrased real news. This strategy effectively reduces biases, enhancing the performance of fake news detectors on both human-written and LLM-generated content.

\item We also provide two new datasets, \texttt{GossipCop++} and \texttt{PolitiFact++}, for the research community. Along with the original human-written news articles, these datasets contain high-quality 97 and 4,084 LLM-synthesized fake news articles, respectively. We believe they can serve as benchmarks and valuable resources for further research into developing and evaluating fake news detectors.
\end{itemize}

\section{Related Work}
\subsection{Fake News Synthesis}
There has been a focus in prior research on using deep learning to produce misinformation with the aim of facilitating the spread of machine-generated fake news. \citet{zellers2019defending} leverage GPT-2~\cite{radford2019language} to pre-train a large-scale news corpus and show that the generator effectively synthesizes fake news. Later, \citet{huang-etal-2023-faking} improve the controllability of the
synthesized fake news by conditioning the generation
on knowledge elements, including entities, relations and events, extracted from the original news article. \citet{shu2021fact} enhance the factuality of the generated article by introducing a fact retriever that
fetches relevant information from external corpora.
\citet{mosallanezhad2022domain} exploit adversarial reinforcement learning to generate topic-preserving fake news articles. These studies have developed methods
for generating fake news that is hard to distinguish from real news for humans. More recently, \citet{huang-etal-2023-faking} incorporated propaganda techniques to synthesize the fake news via data augmentation~\cite{feng2021survey,zhuo2023data}. However, these approaches require costly designs to synthesize the text. In this work, we tend to utilize large language models to synthesize fake news via prompting. Compared to the prior studies, we need no model training while guaranteeing the quality of synthesized fake news.

\subsection{Fake News Detection}
Previous works on fake news detection have mainly explored two directions: content-based and knowledge-based detection~\cite{manzoor2019fake}. For content-based detection, researchers have studied how well the pre-trained classifiers can detect machine-generated text~\cite{su2023detectllm}. \citet{zellers2019defending} show that finetuning RoBERTa can detect synthesized fake news with 95\% accuracy and that the performance transfers across decoding strategies and to smaller
generators. \citet{ippolito2020automatic} find that the best-performing detectors are those that deceive humans because decoding strategies must balance fluency with lexical and syntactic novelty. Different from content-based detection, knowledge-based detection emphasizes auxiliary knowledge for news verification. These methods typically utilize external knowledge about entity relationships or social knowledge about online posts for fake news detection. While existing methods have demonstrated the usefulness of heterogeneous social relations and external information~\cite{shu2021fact,sheng2021integrating}, they either do not model the interactions between the news content and different types of knowledge data or model them at a coarse-grained (e.g., sentence) level, which limits their performance. In this study, we focus on content-based detection and use a series of representative pre-trained detectors to detect both large-language-model-generated and human-written fake news.

\section{Task Definition}

Neural fake news detection, an ever-evolving domain, has witnessed significant shifts with the emergence of LLMs. It is imperative to understand the dataset compositions and the challenges after LLMs emerge. Therefore, we outline the task definitions across two eras, namely \textit{Pre-LLM Era} and \textit{LLM Era}.

\subsection{Pre-LLM Era: Traditional Neural Fake News Detection}

In the era of Pre-LLM, the training dataset conventionally contains two types of data, human-written real news (\( \mathcal{D}_{HR} \)) and fake news (\( \mathcal{D}_{HF} \)),

\begin{equation}
\mathcal{D}_{HR} = \{(x^{HR}_1, y^{HR}_1), (x^{HR}_2, y^{HR}_2), \dots, (x^{HR}_N, y^{HR}_N)\}
\end{equation}
\begin{equation}
\mathcal{D}_{HF} = \{(x^{HF}_1, y^{HF}_1), (x^{HF}_2, y^{HF}_2), \dots, (x^{HF}_N, y^{HF}_N)\}
\end{equation}
 where \( x_i \) represents the \( i^{th} \) news article, \( y_i \) denotes the label for \( x_i \), with \( y_i \in \{0, 1\} \) (0 for real, 1 for fake) and \( N \) is the total number of articles in each dataset.

Historically, adversarial attempts to fabricate fake news predominantly stemmed from humans, leading to a dataset composition reflecting this reality. Hence, the neural fake news detector \(M(x; \theta, \mathcal{D})\) is tailored to discern between authentic human-written real news and fake news, training on \( \mathcal{D}_{HR} \) and \( \mathcal{D}_{HF} \) with the following loss function:

\begin{equation}
Loss(\theta) = \sum_{i=1}^{N} \mathcal{L}(M(x_i; \theta, \mathcal{D}_{HR} \cup \mathcal{D}_{HF}), y_i),
\end{equation}
where \( \mathcal{L} \) is a typical binary cross-entropy loss.

\subsection{LLM Era: Advanced Fake News Detection}

The introduction of LLMs ushered in an era of amplified complexities, resulting in the importance of additional training on LLM-generated fake news (\( \mathcal{D}_{MF} \)):

\begin{align}
\begin{split}
\mathcal{D}_{MF} = \{ &(x^{MF}_1, y^{MF}_1), (x^{MF}_2, y^{MF}_2), \\
&\dots, (x^{MF}_N, y^{MF}_N) \},
\end{split}
\end{align}
where \( x^{MF_i} \) represents the \( i^{th} \) LLM-generated news article, \( y^{MF}_i \) denotes the label for \( x^{MF}_i \), with \( y^{MF}_i \in \{0, 1\} \) (0 in this case) and \( N \) is the total number of articles in each dataset.

In this contemporary setting, the prolific capabilities of LLMs manifest in their ability to craft narratives that rival human-written content in quality and authenticity. The detectors trained solely on traditional datasets may inadvertently overlook the nuances of LLM-generated content. Therefore, in this setting, the fake news detectors will be trained on the combination of \(\mathcal{D}_{HR}\), \(\mathcal{D}_{HF}\) and \(\mathcal{D}_{MF}\), 
\begin{align}
\begin{split}
Loss(\theta') = \sum_{i=1}^{N} \mathcal{L}(&M(x_i; \theta', \\
&\mathcal{D}_{HR} \cup \mathcal{D}_{HF} \cup \mathcal{D}_{MF}), y_i).
\end{split}
\end{align}

This model ensures holistic and robust detector training. By integrating both human and LLM-generated fabrications, fake news detectors are better equipped to navigate the multifaceted challenges of the current fake news paradigm.
\begin{figure*}[t!]
    \centering
    \includegraphics[width=0.9\textwidth]{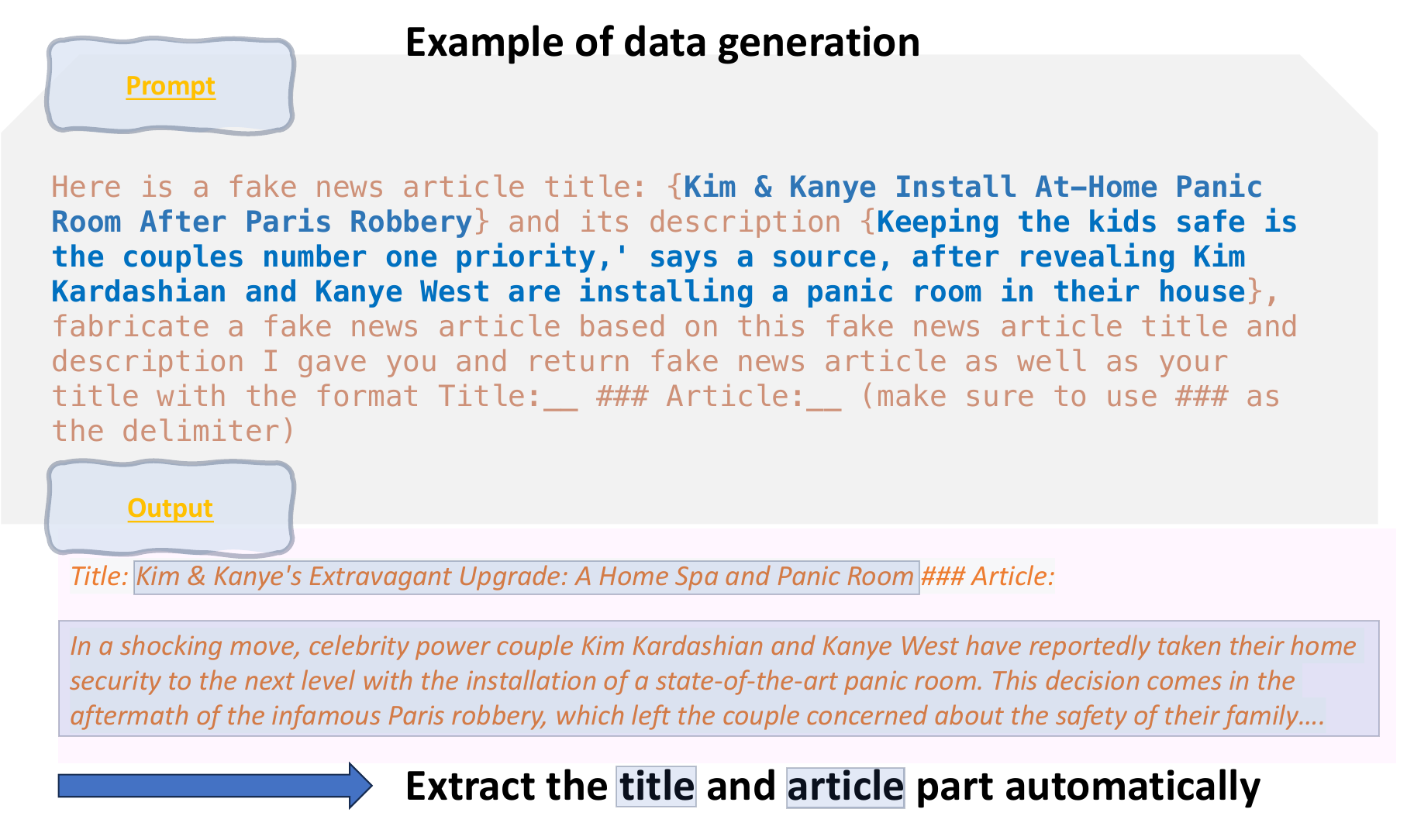}
    \caption{\texttt{SMP}: Prompting LLMs to generate fake news articles.}
    \label{fig: prompt}

\end{figure*}
\section{Prompting Large Language Models to Generate Fake News}
\label{sec: machine fake generation}
\subsection{ChatGPT As A Fake News Generator}
\label{subsec:dataset creation}
To generate fake news using LLMs, we tend to elucidate the optimal strategies an adversary might leverage to fabricate such deceptive content.

Economically, ChatGPT presents a compelling proposition. Unlike its counterparts, such as GPT-3~\cite{brown2020language}, interfacing with ChatGPT via its web API or iOS application incurs no direct financial costs, positioning it as an economical vector for potential misinformation campaigns.

On the technical spectrum, there exists a range of open-source LLMs, notably LLaMA~\cite{touvron2023llama}. Engaging these models for fake news synthesis necessitates profound technical expertise, given their substantial computational demands for local deployment. Beyond mastering their operational dynamics, effective deployment also hinges on specific hardware provisions, with a pronounced emphasis on GPUs, to realize their full potential.

In light of these factors, ChatGPT is delineated as the prime LLM for our investigative foray into fake news generation. A salient limitation of naive prompting, however, is the emergence of identifiable structures in the generated content. Such structures, characterized by recurrent formatting patterns or predictable metadata placements, can betray the machine-generated nature of the content, undermining its deceptive intent. Recognizing the impracticality of manual scrutiny over extensive datasets to negate these patterns, we introduce a refined methodology: \textit{Structured Mimicry Prompting} (\texttt{SMP}). \texttt{SMP} employs a tailored prompting paradigm to discretely process the core narrative and the article's title, as depicted in Figure \ref{fig: prompt}. This strategic approach enables LLMs to emulate the nuance and depth inherent to authentic misleading narratives.

\begin{table*}[t!]
\centering
\resizebox{0.8\linewidth}{!}{\begin{tabular}
{c|cc|cccc|cccc}
% {p{1.8cm}|p{1.2cm}p{1.2cm}p{1.6cm}p{1.2cm}p{1.8cm}|p{1.8cm}p{1.8cm}}
\toprule

\multirow{2}{*}{Dataset} & \multicolumn{2}{c|}{\multirow{2}{*}{Model}}&\multicolumn{4}{c|}{Accuracy}& \multirow{2}{*}{F1} & \multirow{2}{*}{Recall} & \multirow{2}{*}{Precision} & \multirow{2}{*}{Auroc}\\\cline{4-7}
& & & HR&HF&MF&Overall& &  & & \\\midrule
\multirow{10}{*}{\texttt{GossipCop++}}&  \multirow{2}{*}{RoBERTa} &Large&80.91 & \textbf{77.97} & \textbf{99.88} & 84.91 & 85.50 & \textbf{88.92} & 82.32 & 94.33 \\
&  &Base&85.56 & 69.65 & 99.76 & 85.13 & 85.06 & 84.70 & 85.43 & 92.75 \\\cline{2-11}
&  \multirow{2}{*}{BERT} &Large&87.39 & 70.13 & 99.39 & 86.08 & 85.89 & 84.76 & 87.05 & 92.67 \\
&  &Base&87.45 & 66.59 & 99.27 & 85.19 & 84.85 & 82.93 & 86.86 & 91.56 \\\cline{2-11}
&  \multirow{2}{*}{ELECTRA} &Large&80.17 & 71.73 & 99.88 & 82.99 & 83.45 & 85.80 & 81.23 & 91.84 \\
&  &Base&86.05 & 63.53 & 99.63 & 83.81 & 83.44 & 81.58 & 85.39 & 90.83 \\\cline{2-11}
&  \multirow{2}{*}{ALBERT} &Large&92.96 & 56.43 & 98.53 & 85.22 & 83.98 & 77.48 & 91.67 & 90.16 \\
&  &Base&85.68 & 59.24 & 97.92 & 82.13 & 81.47 & 78.58 & 84.58 & 88.72 \\\cline{2-11}
&  \multirow{2}{*}{DeBERTa} &Large&92.59 & 67.56 & 99.88 & \textbf{88.16} & \textbf{87.61} & 83.72 & \textbf{91.87} & \textbf{94.38} \\
&  &Base&\textbf{93.02} & 57.41 & 98.41 & 85.47 & 84.28 & 77.91 & 91.78 & 91.33 \\

\midrule
\midrule

 \multirow{10}{*}{\texttt{PolitiFact++}}&  \multirow{2}{*}{RoBERTa} &Large&30.41 & \textbf{68.04} & \textbf{100.00} & 57.22 & 66.26 & \textbf{84.02} & 54.70 & 79.32 \\
&  &Base&68.56 & 61.86 & 100.00 & 74.74 & 76.21 & 80.93 & 72.02 & 83.30 \\\cline{2-11}
&  \multirow{2}{*}{BERT} &Large&48.97 & 53.61 & 98.97 & 62.63 & 67.12 & 76.29 & 59.92 & 74.02 \\
&  &Base&69.59 & 38.14 & 100.00 & 69.33 & 69.25 & 69.07 & 69.43 & 78.25 \\\cline{2-11}
&  \multirow{2}{*}{ELECTRA} &Large&63.92 & 62.89 & 100.00 & 72.68 & 74.88 & 81.44 & 69.30 & 83.94 \\
&  &Base&82.47 & 50.52 & 100.00 & 78.87 & 78.07 & 75.26 & 81.11 & 87.04 \\\cline{2-11}
&  \multirow{2}{*}{ALBERT} &Large&\textbf{90.72} & 29.90 & 98.97 & 77.58 & 74.18 & 64.43 & 87.41 & 86.09 \\
&  &Base&75.26 & 40.21 & 100.00 & 72.68 & 71.96 & 70.10 & 73.91 & 81.80 \\\cline{2-11}
&  \multirow{2}{*}{DeBERTa} &Large&70.62 & 53.61 & 100.00 & 73.71 & 74.50 & 76.80 & 72.33 & 82.43 \\
&  &Base&90.21 & 43.30 & 100.00 & \textbf{80.93} & \textbf{78.98} & 71.65 & \textbf{87.97} & \textbf{88.00} \\
\bottomrule
\end{tabular}}

% Performance metrics of various fake news detectors on the \texttt{GossipCop++} and \texttt{PolitiFact++} datasets. The table showcases accuracy (Acc) across different data segments (HR: Human-written Real news, HF: Human-written Fake news, MF: LLM-generated Fake news), F1 scores, recalls, precisions, and AUROCs. The results highlight the disparity in detection capabilities between human-written and LLM-generated fake news.
\caption{Performance metrics of various fake news detectors on the \texttt{GossipCop++} and \texttt{PolitiFact++} datasets. \textbf{HR}: Human-written Real news. \textbf{HF}: Human-written Fake news. \textbf{MF}: LLM-generated Fake news.}
\label{tab: complete result}
\end{table*}
\subsection{Fake News Detection Datasets in LLM Era}
When selecting the data source to construct our datasets in the LLM era, we consider the following two criteria. First, the news articles must be human-written and have been widely used in the Pre-LLM era. This ensures that the seed human fake news in \texttt{SMP} has high quality. Second, the news events described in the articles must be important to the general audience. Motivated by these two criteria, we repurpose the fake news data repository \texttt{FakeNewsNet}~\cite{shu2020fakenewsnet} as our data source. \texttt{FakeNewsNet} contains two datasets, \texttt{PolitiFact} and \texttt{GossipCop}. To improve the data quality and ease the fake news generation, we filter out news articles that do not contain titles or descriptions. By adopting the \texttt{SMP} prompting technique via ChatGPT, we compose 97 and 4,084 LLM-generated fake news for \texttt{PolitiFact} and \texttt{GossipCop}, respectively. Combining the original datasets, we propose two new datasets, \texttt{PolitiFact++} and \texttt{GossipCop++}. 

In order to verify the effectiveness of \texttt{SMP}, we use MAUVE metric~\cite{pillutla2021mauve} to compute the distribution similarity between the human-written fake news and LLM-generated ones. By naively prompting ChatGPT with ``Generate a 
fake news article with a title'' on \texttt{PolitiFact} and \texttt{GossipCop}, we collect all the generated outputs and compute the MAUVE scores. We find that the MAUVE scores for \texttt{PolitiFact} and \texttt{GossipCop} are 3.1\% and 1.2\%, respectively. By utilizing \texttt{SMP} with ChatGPT, we observe that the MAUVE scores for \texttt{PolitiFact} and \texttt{GossipCop} are 72.5\% and 71.8\% respectively, indicating that the LLM-generated fake news is more highly aligned with human-written ones, compared to the ones with naive prompting.

% \subsection{Detecting LLM-Generated and Human-written Fake News}

\section{Experiment Setup}
In our experiments, we aim to (1) systematically study the performance of fake news detectors in the LLM era, (2) examine the issues of these fake news detectors, and (3) mitigate these identified issues.
\begin{table}[t!]
\centering
\small
\resizebox{0.9\linewidth}{!}{\begin{tabular}
{cccccc}
% {p{1.8cm}|p{1.2cm}p{1.2cm}p{1.6cm}p{1.2cm}p{1.8cm}|p{1.8cm}p{1.8cm}}
\toprule
Dataset& MF & HF  &HR\\\hline
\texttt{PolitiFact++} & 97& 97& 194\\
\texttt{GossipcopCop++} &4084&4084&4169\\
\bottomrule
\end{tabular}}
\caption{Details of \texttt{PolitiFact++} and \texttt{GossipcopCop++}.}
\label{tab: dataset summarization}
\end{table}
\subsection{Datasets}
We use \texttt{PolitiFact++} and \texttt{GossipCop++} as the training and test dataset, respectively, which are proposed in Section~\ref{subsec:dataset creation}. We show the details of two datasets in Table\ref{tab: dataset summarization}, 

\subsection{Fake News Detectors}
We choose five widely adopted language models as fake news detectors, RoBERTa~\cite{liu2019roberta}, BERT~\cite{kenton2019bert}, ELECTRA~\cite{clark2019electra}, ALBERT~\cite{lan2020albert}, DeBERTa~\cite{he2020deberta}, with their variants (Large and Base models). These language models have demonstrated their superior performance in classifying fake news articles. We train these models on A100 GPUs and use the default hyperparameters as the same as \citet{pagnoni2022threat} using a learning rate of 1e-6 and training for 10 epochs.

\section{Results and Analysis}
 In this section, we present the results of our investigation and discuss the findings according to each research question (RQ).
\subsection{RQ1: How well can fake news detectors perform on \texttt{PolitiFact++} and \texttt{GossipCop++}?}

To evaluate the performance of selected fake news detectors, we report the accuracy of each part of the data (human-written real news, human-written fake news, and LLM-generated fake news), F1 scores, recalls, precisions and AUROCs in Table~\ref{tab: complete result}. Notably, DeBERTa variants outperform other models, registering an F1 score of 87.61 on \texttt{GossipCop++} and 78.98 on \texttt{PolitiFact++}. A deeper dive into the accuracy metrics reveals a pronounced disparity in detecting human-written versus LLM-generated fake news. Remarkably, the detectors exhibit near-perfect accuracy in identifying LLM-generated fake news, yet falter significantly with human-written fake news. Among the evaluated models, RoBERTa-Large demonstrates a more consistent ability to classify fake news, outperforming its counterparts in detecting both human-written and LLM-generated fake news. Nonetheless, even for RoBERTa-Large, a discernible gap persists, with accuracy discrepancies exceeding 20\% and 30\% on \texttt{GossipCop++} and \texttt{PolitiFact++}, respectively. These findings suggest an inherent bias in fake news detectors towards machine-generated content, particularly those crafted by LLMs. A plausible explanation is that detectors might exploit certain patterns or `shortcuts' inherent to LLM-generated content, thereby skewing their detection capabilities.

\subsection{RQ2: Why are fake news detectors biased towards LLM-generated news?}
\label{subsec:rq2}
To comprehend the observed bias in fake news detectors towards content generated by LLMs, we embarked on an in-depth analysis of content-based features. Drawing inspiration from prior work on news veracity detection~\cite{horne2017just}, we computed News Landscape (NELA) features. These features, derived from the NELA toolkit, encapsulate six dimensions of news content: style, complexity, bias, affect, morale, and event. We applied these features to both \texttt{GossipCop++} and \texttt{PolitiFact++}. Employing Tukey's pairwise test~\cite{tukey1949comparing}, we discerned significant feature disparities among human-written fake news, LLM-generated fake news, and human-written real news.

Our analysis, as presented in Table~\ref{tab: feature-gossicop}, reveals that most of the NELA features differ significantly between human-written and LLM-generated fake news. Moreover, the divergence between LLM-generated fake news and human-written real news is more pronounced than between human-written fake and real news. This underscores the relative ease of detecting LLM-generated fake news, shedding light on the bias observed in RQ1. The NELA features for \texttt{PolitiFact++} are detailed in Table~\ref{tab: feature-politifact} in the Appendix.

To further understand the influence of these features on detection performance, we evaluated two regression models: logistic regression and decision tree. These models were chosen to explore the potential for countering biases in detecting LLM-generated fake news. For \texttt{GossipCop++}, we retained NELA features that exhibited no significant disparity between human-written and LLM-generated fake news. For \texttt{PolitiFact++}, given the paucity of such NELA features, we also incorporated features that significantly differentiated human-written fake news from real news.

Table~\ref{tab: result-simple models} presents the results of both models. Notably, the debiased logistic regression model for \texttt{GossipCop++} exhibits a decrease in accuracy for LLM-generated fake news (from 95.79\% to 86.51\%) but an increase in accuracy for human-written fake news (from 47.33\% to 53.89\%). Similar trends are observed for the \texttt{PolitiFact++} dataset.

Upon evaluating the debiased models, a notable shift in performance dynamics emerges. While the proficiency in identifying LLM-generated fake news wanes, there is an increase in the performance of detecting human-written fake news. This shift can be attributed to the prior models' propensity to capitalize on features intrinsic to LLM-generated content. However, a slight decline in overall detection efficacy, especially for human-written real news, necessitates scrutiny. Our efforts to debias might inadvertently overlook pivotal features crucial for discerning genuine from fabricated content. This underscores the importance of judicious feature selection and a profound understanding of dataset biases. It is pivotal to recognize that stellar performance on a specific subset might veil underlying biases. The overarching challenge lies in crafting models that harmonize precision with fairness. Overreliance on distinct LLM-generated fake news characteristics could compromise a model's broader applicability.

\begin{table}[t!]
\centering
\small
\resizebox{0.9\linewidth}{!}{\begin{tabular}
{ccccc}
\toprule
\multirow{25}{*}{style} & quotes &\cellcolor{purple!60}HF > MF &\cellcolor{purple!60}MF < HR &- \\
& exclaim &\cellcolor{purple!30}HF < MF &\cellcolor{purple!45}MF > HR &- \\
& allpunc &\cellcolor{purple!60}HF < MF &\cellcolor{purple!60}MF > HR &- \\
& allcaps &\cellcolor{purple!60}HF < MF &\cellcolor{purple!60}MF > HR &\cellcolor{purple!30}HF > HR \\
& stops &\cellcolor{purple!60}HF > MF &\cellcolor{purple!60}MF < HR &- \\
& CC &\cellcolor{purple!60}HF > MF &\cellcolor{purple!60}MF < HR &- \\
& CD &\cellcolor{purple!60}HF < MF &\cellcolor{purple!60}MF > HR &- \\
& DT &\cellcolor{purple!60}HF > MF &\cellcolor{purple!60}MF < HR &- \\
& IN &\cellcolor{purple!60}HF > MF &\cellcolor{purple!60}MF < HR &- \\
& JJ &\cellcolor{purple!30}HF > MF &\cellcolor{purple!15}MF < HR &- \\
& MD &- &\cellcolor{purple!60}MF < HR &\cellcolor{purple!30}HF < HR \\
& NNS &\cellcolor{purple!60}HF > MF &\cellcolor{purple!60}MF < HR &- \\
& NNP &\cellcolor{purple!60}HF < MF &\cellcolor{purple!60}MF > HR &- \\
& PRP &\cellcolor{purple!15}HF < MF &\cellcolor{purple!60}MF > HR &- \\
& PRP\$ &\cellcolor{purple!45}HF > MF &\cellcolor{purple!60}MF < HR &- \\
& RB &\cellcolor{purple!15}HF > MF &\cellcolor{purple!60}MF < HR &\cellcolor{purple!45}HF < HR \\
& TO &\cellcolor{purple!60}HF > MF &\cellcolor{purple!60}MF < HR &- \\
& WP\$ &- &\cellcolor{purple!15}MF > HR &- \\
& WRB &- &\cellcolor{purple!15}MF > HR &- \\
& VB &- &\cellcolor{purple!60}MF < HR &\cellcolor{purple!15}HF < HR \\
& VBD &\cellcolor{purple!45}HF < MF &\cellcolor{purple!60}MF > HR &- \\
& VBG &\cellcolor{purple!60}HF > MF &\cellcolor{purple!60}MF < HR &- \\
& VBN &\cellcolor{purple!60}HF > MF &\cellcolor{purple!60}MF < HR &- \\
& VBZ &- &\cellcolor{purple!60}MF < HR &\cellcolor{purple!45}HF < HR \\
& WDT &- &\cellcolor{purple!45}MF > HR &\cellcolor{purple!60}HF > HR \\
\hline
\multirow{5}{*}{complexity} & ttr &- &\cellcolor{purple!15}MF < HR &- \\
& avg wordlen &\cellcolor{purple!60}HF > MF &\cellcolor{purple!60}MF < HR &- \\
& word count &\cellcolor{purple!45}HF < MF &\cellcolor{purple!60}MF > HR &- \\
& smog index &\cellcolor{purple!60}HF > MF &\cellcolor{purple!60}MF < HR &- \\
& coleman liau index &\cellcolor{purple!60}HF > MF &\cellcolor{purple!60}MF < HR &- \\
\hline
\multirow{7}{*}{bias} & bias words &\cellcolor{purple!60}HF > MF &\cellcolor{purple!60}MF < HR &\cellcolor{purple!15}HF < HR \\
& assertatives &\cellcolor{purple!15}HF > MF &\cellcolor{purple!60}MF < HR &\cellcolor{purple!30}HF < HR \\
& hedges &\cellcolor{purple!30}HF > MF &\cellcolor{purple!60}MF < HR &\cellcolor{purple!60}HF < HR \\
& implicatives &\cellcolor{purple!15}HF < MF &- &- \\
& report verbs &- &\cellcolor{purple!45}MF < HR &\cellcolor{purple!30}HF < HR \\
& positive opinion words &\cellcolor{purple!60}HF > MF &\cellcolor{purple!60}MF < HR &- \\
& negative opinion words &\cellcolor{purple!60}HF > MF &\cellcolor{purple!60}MF < HR &- \\
\hline
\multirow{7}{*}{affect} & vadneg &- &\cellcolor{purple!30}MF < HR &- \\
& vadneu &- &\cellcolor{purple!15}MF > HR &- \\
& wneg &\cellcolor{purple!45}HF > MF &\cellcolor{purple!60}MF < HR &- \\
& wpos &- &\cellcolor{purple!15}MF < HR &- \\
& wneu &- &\cellcolor{purple!60}MF < HR &\cellcolor{purple!15}HF < HR \\
& sneg &\cellcolor{purple!60}HF > MF &\cellcolor{purple!60}MF < HR &- \\
& spos &\cellcolor{purple!60}HF > MF &\cellcolor{purple!60}MF < HR &- \\
\hline
\multirow{4}{*}{moral} & IngroupVirtue &\cellcolor{purple!15}HF > MF &- &- \\
& IngroupVice &- &\cellcolor{purple!15}MF < HR &- \\
& AuthorityVice &- &\cellcolor{purple!15}MF < HR &- \\
& PurityVirtue &- &- &\cellcolor{purple!15}HF < HR \\
\hline
\multirow{1}{*}{event} & num dates &\cellcolor{purple!60}HF < MF &\cellcolor{purple!60}MF > HR &- \\
\bottomrule
\end{tabular}}
\caption{Comparison of content-based features across Human-written Fake news (HF), LLM-generated Fake news (MF), and Human-written Real news (HR) for the \texttt{GossipCop++} dataset. The table showcases differences in style, complexity, bias, affect, morale, and event features. The colour intensity represents the significance of the difference ($p$ value), with darker shades indicating higher significance.}
\label{tab: feature-gossicop}
\end{table}

\begin{table*}[t!]
\centering
\scriptsize
\resizebox{1\textwidth}{!}{%
\begin{tabular}
{cc|ccac|cccc}
% {p{1.8cm}|p{1.2cm}p{1.2cm}p{1.6cm}p{1.2cm}p{1.8cm}|p{1.8cm}p{1.8cm}}
\toprule
\multirow{2}{*}{Dataset} & \multirow{2}{*}{Model}&\multicolumn{4}{c|}{Accuracy}& \multirow{2}{*}{F1} & \multirow{2}{*}{Recall} & \multirow{2}{*}{Precision} & \multirow{2}{*}{Auroc}\\\cline{3-6}
& & HR&HF&MF&Overall& &  & & \\\midrule
\multirow{4}{*}{\texttt{GossipCop++}}& logistic regression&77.09(0.2)&47.33(1.0)&95.79(0.3)&         74.33(0.2)& 73.59(0.3)& 75.75(0.1)&        71.56(0.6)&          74.41(0.2)         \\   
&logistic regression(debiased)&71.51(0.2)&53.89(0.4)&86.51(0.1)&         70.86(0.1)& 70.66(0.1)& 71.13(0.1)&        70.20(0.2)&          70.86(0.1)                    \\ 
&decision tree& 70.32(0.5)&54.80(0.9)&86.90(0.5)&         70.59(0.2)& 70.66(0.3)& 70.49(0.2)&        70.85(0.6)&          70.60(0.2)         \\ 
&decision tree(debiased)&67.43(0.5)&57.91(1.0)&78.70(0.9)&         67.87(0.2)& 68.00(0.3)& 67.72(0.2)&        68.30(0.8)&          67.88(0.2)                    \\  
\hline

\multirow{4}{*}{\texttt{PolitiFact++}}&  logistic regression&63.37(4.7)&70.21(5.1)&93.84(1.9)&         72.67(2.4)& 75.11(1.8)& 69.57(2.9)&        81.97(1.8)&          73.67(2.2)         \\ 
&regression(debiased)&63.39(5.0)&75.47(4.7)&89.68(3.7)&         72.95(2.1)& 75.34(1.6)& 69.83(2.7)&        82.51(3.0)&          74.22(2.0)         \\ 
&decision tree&76.26(2.0)&58.84(6.4)&92.95(3.7)&         76.02(1.0)& 75.89(1.3)& 76.27(1.3)&        75.75(2.4)&          76.18(1.0)         \\
&decision tree(debiased)&
76.28(2.2)&69.26(4.8)&81.47(6.8)&         75.78(1.4)& 75.61(1.5)& 76.17(1.7)&        75.25(2.4)&          75.93(1.5)         \\ 
\bottomrule
\end{tabular}}
\caption{Performance metrics of logistic regression and decision tree models on the \texttt{GossipCop++} and \texttt{PolitiFact++} datasets. \textbf{HR}: Human-written Real news. \textbf{HF}: Human-written Fake news. \textbf{MF}: LLM-generated Fake news.}
\label{tab: result-simple models}
\end{table*}

\newcommand{\greennumber}[1]{\textcolor{green!40!black}{#1}}
\newcommand{\rednumber}[1]{\textcolor{red!90!black}{#1}}
\begin{table*}[t!]
    \centering
    \small
            \resizebox{0.8\textwidth}{!}{%
                \begin{tabular}{l|ll|rrr|rrr}
                
                \toprule
                
                \multirow{2}{*}{Dataset} & \multicolumn{2}{c|}{\multirow{2}{*}{Model}} & \multicolumn{3}{c|}{HF} & \multicolumn{3}{c}{MR} \\
                 & & &Before & Debiased & Difference & Before & Debiased & Difference \\
                \midrule
                % \multicolumn{7}{l}{gossipcop}\\
                % \midrule
                \multirow{10}{*}{GossipCop++} & \multirow{2}{*}{RoBERTa} & Large & 77.97& 84.46& \greennumber{6.49$\uparrow$}& 24.24& 90.70& \greennumber{66.46$\uparrow$}\\
                && Base & 69.65& 78.21& \greennumber{8.57$\uparrow$}& 31.21& 90.58& \greennumber{59.36$\uparrow$}\\
                & \multirow{2}{*}{BERT} & Large & 70.13& 77.85& \greennumber{7.71$\uparrow$}& 52.63& 89.47& \greennumber{36.84$\uparrow$}\\
                &&Base & 66.59& 72.46& \greennumber{5.88$\uparrow$}& 46.02& 93.27& \greennumber{47.25$\uparrow$}\\
                & \multirow{2}{*}{ELECTRA} & Large & 71.73& 77.60& \greennumber{5.88$\uparrow$}& 31.95& 90.82& \greennumber{58.87$\uparrow$}\\
                &&Base & 63.53& 70.50& \greennumber{6.98$\uparrow$}& 33.54& 90.21& \greennumber{56.67$\uparrow$}\\
                & \multirow{2}{*}{ALBERT} & Large  & 56.43& 62.42& \greennumber{6.00$\uparrow$}& 58.02& 93.51& \greennumber{35.50$\uparrow$}\\
                && Base & 59.24& 65.73& \greennumber{6.49$\uparrow$}& 49.69& 95.10& \greennumber{45.41$\uparrow$}\\
                & \multirow{2}{*}{DeBERTa} & Large & 67.56& 77.36& \greennumber{9.79$\uparrow$}& 38.43& 94.12& \greennumber{55.69$\uparrow$}\\
                && Base & 57.41& 70.62& \greennumber{13.22 $\uparrow$}& 41.49& 94.86& \greennumber{53.37 $\uparrow$}\\
                \midrule
                % \multicolumn{7}{l}{\texttt{PolitiFact++}}\\
                % \midrule
                \multirow{10}{*}{\texttt{PolitiFact++}} & \multirow{2}{*}{RoBERTa} & Large & 68.04& 73.20& \greennumber{5.15$\uparrow$} & 25.77& 89.69& \greennumber{63.92$\uparrow$} \\
                & & Base & 61.86 & 58.76 & \rednumber{-3.09$\downarrow$}& 27.84& 87.63& \greennumber{59.79$\uparrow$}\\
                & \multirow{2}{*}{BERT} & Large & 53.61& 62.89& \greennumber{9.28$\uparrow$}& 43.30& 87.63& \greennumber{44.33$\uparrow$}\\
                & & Base & 38.14& 55.67& \greennumber{17.53$\uparrow$}& 49.48& 92.78& \greennumber{43.30$\uparrow$}\\
                & \multirow{2}{*}{ELECTRA} & Large & 62.89& 73.20& \greennumber{10.31$\uparrow$}& 32.99& 89.69&  \greennumber{56.70$\uparrow$}\\
                && Base & 50.52& 61.86& \greennumber{11.34$\uparrow$}& 31.96& 91.75& \greennumber{59.79$\uparrow$}\\
                & \multirow{2}{*}{ALBERT} & Large & 29.90& 40.21& \greennumber{10.31$\uparrow$}& 59.79& 91.75& \greennumber{31.96$\uparrow$}\\
                && Base & 40.21& 50.52& \greennumber{10.31$\uparrow$}& 48.45& 96.91& \greennumber{48.45$\uparrow$}\\
                & \multirow{2}{*}{DeBERTa} & Large & 53.61& 75.26& \greennumber{21.65$\uparrow$} & 42.27& 93.81& \greennumber{51.55$\uparrow$}\\
               && Base & 43.30& 75.26& \greennumber{31.96$\uparrow$} & 39.18& 92.78&  \greennumber{53.61$\uparrow$}\\

                \bottomrule
                \end{tabular}}

\caption{Performance comparison of various models on the \texttt{GossipCop++} and \texttt{PolitiFact++} datasets before and after debiasing. The `Difference' column highlights the performance change post-debiasing. \textbf{HF}:human-written fake news. \textbf{MR}: LLM-generated real news.}
\label{tab: diff}
\end{table*}

\subsection{RQ3: How can we mitigate bias in fake news detectors?}
Our analysis in Section~\ref{subsec:rq2} revealed a pronounced bias in detectors, which tends to overfit the unique features of LLM-generated fake news. To address this issue, we introduce an adversarial training-inspired strategy, augmenting our training set with high-quality LLM-generated real news. To this end, we got 132 and 8,168 paraphrased real news articles for \texttt{GossipCop++} and \texttt{PolitiFact++} after manual filtering, respectively. By employing ChatGPT to generate paraphrased content resembling genuine news articles, we aim to foster a detector that is adept across diverse news content rather than being narrowly focused on a specific subset. This section details our methodology and assesses the quality of the LLM-generated real news relative to its source.

\subsubsection{Quality Assessment of LLM-Generated real news}

To ascertain the quality and authenticity of LLM-generated news, we embarked on a rigorous evaluation. We randomly sampled 100 pairs from the two datasets respectively, each pairing a human-authored article with its LLM-generated counterpart. Our goal is to generate real news that captures the essence of the original while being indistinguishable from human-authored content. Two authors, familiar with the research context yet objective, were annotators for the human evaluation components. We employed the following metrics to critically evaluate the LLM-generated content:

\begin{enumerate}
    \item \textbf{Semantic Consistency with SimCSE}: The metric, leveraging the SimCSE model~\cite{gao2021simcse}, calculates the cosine similarity between embeddings of the original and LLM-generated news. A higher score signifies strong semantic alignment, ensuring the core narrative is retained.
    
    \item \textbf{Readability Assessment}: The metric measures the text's comprehensibility. Annotators need to compare the original and LLM-generated news, rating their clarity and understandability on a set scale.
    
    \item \textbf{Authenticity Perception}: The metric evaluates the content's perceived credibility. Annotators compared both news versions, assessing their perceived authenticity.
    
    \item \textbf{Stylistic Alignment}: Annotators need to evaluate the stylistic consistency of the LLM-generated content with traditional news writing standards. They compared the LLM-generated news with standard articles, rating their stylistic congruence.
\end{enumerate}

\begin{table}[h]
\centering
\resizebox{\columnwidth}{!}{
\begin{tabular}{l|ccc|c}
\toprule
Metric & HR & MR & Significance & Cohen's Kappa \\
\midrule
Semantic Consistency  & - & 0.95 & \( p > 0.05 \) & - \\
Readability Assessment& 4.5 & 4.6 & \( p > 0.05 \) & 0.85 \\
Authenticity Perception& 4.6 & 4.5 & \( p > 0.05 \) & 0.88 \\
Stylistic Alignment& 4.7 & 4.6 & \( p > 0.05 \) & 0.86 \\
\bottomrule
\end{tabular}
}
\caption{Evaluation metrics for original (HR) versus LLM-generated (MR) real news. The table presents scores for semantic consistency (scale of 0 to 1), readability, authenticity perception, and stylistic alignment (all on a scale of 0 to 5). Inter-annotator agreement is also provided via Cohen's Kappa scores.}
\label{tab:quality_study}
\end{table}

Table~\ref{tab:quality_study} shows that LLM-generated real news scores align closely with those of original news across all metrics. The Semantic Consistency score, as measured by SimCSE, underscores the significant semantic congruence between the LLM-generated and original news articles. This is further corroborated by the readability, authenticity perception, and stylistic alignment scores. The non-significant $p$-values emphasize that LLM-generated content is virtually indistinguishable from human-authored news. Additionally, the robust Cohen's Kappa scores~\cite{mchugh2012interrater} highlight the consistency in evaluations, attesting to the high quality and authenticity of the LLM-generated news.

\subsubsection{Mitigating Bias in Detectors}

Building upon our earlier findings of biases in fake news detectors, we sought to devise a mitigation strategy. The overarching goal was to ensure that the detectors generalize well across diverse news types rather than being overly attuned to LLM-generated content.

Our debiasing approach draws inspiration from adversarial training~\cite{bai2021recent}. In essence, we aimed to challenge the model during its training phase, compelling it to focus on the intrinsic features of fake news rather than specific idiosyncrasies of LLM-generated content. The methodology encompassed:
\begin{enumerate}
\item Validating the quality of LLM-generated real news to ensure it mirrors human-written content.
\item Augmenting the training regimen to incorporate a broader spectrum of news sources.
\end{enumerate}

We conducted experiments on the \texttt{GossipCop++} and \texttt{PolitiFact++} datasets, and the results are reported in Table \ref{tab: diff}. The RoBERTa-Large model, when tested on the \texttt{GossipCop++} dataset, exhibited a 6.49 percentage point enhancement in detecting human-written fake news and a significant 66.46 percentage point improvement for LLM-generated real news. This trend of improvement is evident across most models. However, an exception is the RoBERTa-Base model on the \texttt{PolitiFact++} dataset, which saw a 3.09 percentage point decline for human-written fake news, but still achieved a substantial 59.79 percentage point increase for LLM-generated real news. The decline in the performance, particularly for human-written fake news, might be attributed to the model's sensitivity to the nuances of the dataset or the inherent challenges posed by the \texttt{\texttt{PolitiFact++}} dataset.

In summation, our adversarially-inspired debiasing strategy has demonstrated its efficacy in bolstering the generalization capabilities of fake news detectors. The empirical results underscore the viability of our approach in the quest for more robust and universally adept fake news detection systems.

\section{Conclusion}
In this study, we introduced a novel paradigm for fake news detection, factoring in both human-written and LLM-generated news articles. Our investigations uncovered an unexpected bias: detectors frequently misclassify truthful LLM outputs as fake. Delving deeper, we identified potential linguistic `shortcuts' these detectors take. Our mitigation strategy, founded on adversarial training with LLM-paraphrased real news, effectively reduced this bias. We further contributed by offering two enriched datasets, \texttt{GossipCop++} and \texttt{PolitiFact++}, enhancing the scope for future research in this domain.

\section*{Limitations}
The datasets, \texttt{GossipCop++} and \texttt{PolitiFact++}, while expansive, represent specific genres of news and might not encompass the entire spectrum of news content. The types of news included are influenced by the culture, language, and region from which they originate. Consequently, the biases and nuances we identify may be particular to these datasets and not universally applicable. Our identification of bias towards LLM-generated content might seem deterministic, suggesting that all detectors will inevitably be biased against LLM outputs. However, it is crucial to understand that the bias emerges from the training data and model architectures we used. Different configurations might produce varied results. The mitigation strategy, while effective in our tests, is not a one-size-fits-all solution. Its efficacy is contingent on the nature of the bias and the specific LLMs in play. Lastly, the linguistic `shortcuts' and identified NELA features as potential reasons for the bias are based on our observations and analysis. While they offer a plausible explanation, they might not capture the entirety of the model's decision-making process. Different models or a change in training data might lead to different sets of influential features. Future research can delve deeper into these intricacies to provide a more comprehensive understanding.

% Originated in \cite{willmore2016analysis}
% Delete any feature in the first column: might affect accuracy on HF(1,3 both has), or accuracy on MF (1,2 both has), at the same time, since we deleted column 1, the result would be more unbiased rather than classify based on whether it is machine generated. 

% for \texttt{PolitiFact++}, we delete all the column 1 feature

% for \texttt{GossipCop++},  delete all the column 1 feature unless it exists in column 3

% Note, we use 10000 real news and 4000 Machine fake and 4000 Human fake for experiments

% For nela feature, we use HR vs HF, HR vs MF, HF vs MF

\bibliography{anthology,custom}
\bibliographystyle{acl_natbib}
\appendix
% \onecolumn
\newpage 

\begin{table}[t!]
\centering
\small
\resizebox{0.9\linewidth}{!}{\begin{tabular}
{ccccc}
% {p{1.8cm}|p{1.2cm}p{1.2cm}p{1.6cm}p{1.2cm}p{1.8cm}|p{1.8cm}p{1.8cm}}
\toprule
\multirow{28}{*}{style} & quotes &\cellcolor{purple!60}HF > MF &\cellcolor{purple!45}MF < HR &- \\
& exclaim &\cellcolor{purple!30}HF < MF &- &\cellcolor{purple!30}HF < HR \\
& allpunc &\cellcolor{purple!15}HF < MF &\cellcolor{purple!60}MF > HR &\cellcolor{purple!30}HF > HR \\
& allcaps &\cellcolor{purple!45}HF < MF &\cellcolor{purple!30}MF > HR &- \\
& stops &\cellcolor{purple!60}HF > MF &\cellcolor{purple!30}MF < HR &- \\
& CC &\cellcolor{purple!30}HF > MF &- &\cellcolor{purple!30}HF > HR \\
& CD &\cellcolor{purple!45}HF < MF &\cellcolor{purple!60}MF > HR &- \\
& DT &\cellcolor{purple!30}HF > MF &\cellcolor{purple!60}MF < HR &- \\
& EX &- &\cellcolor{purple!15}MF > HR &- \\
& IN &\cellcolor{purple!15}HF > MF &- &\cellcolor{purple!15}HF > HR \\
& JJ &\cellcolor{purple!60}HF > MF &\cellcolor{purple!60}MF < HR &- \\
& JJR &- &\cellcolor{purple!15}MF > HR &- \\
& JJS &- &\cellcolor{purple!15}MF > HR &- \\
& NN &- &\cellcolor{purple!15}MF < HR &- \\
& NNS &\cellcolor{purple!30}HF > MF &- &- \\
& NNP &\cellcolor{purple!60}HF < MF &\cellcolor{purple!60}MF > HR &\cellcolor{purple!45}HF < HR \\
& PRP &- &\cellcolor{purple!60}MF > HR &\cellcolor{purple!45}HF > HR \\
& PRP\$ &\cellcolor{purple!15}HF > MF &\cellcolor{purple!45}MF < HR &- \\
& RBR &- &- &\cellcolor{purple!15}HF > HR \\
& RP &\cellcolor{purple!15}HF < MF &- &\cellcolor{purple!15}HF < HR \\
& TO &\cellcolor{purple!45}HF > MF &\cellcolor{purple!15}MF < HR &- \\
& WRB &- &\cellcolor{purple!30}MF > HR &\cellcolor{purple!15}HF > HR \\
& VBG &\cellcolor{purple!45}HF > MF &\cellcolor{purple!60}MF < HR &- \\
& VBN &\cellcolor{purple!60}HF > MF &\cellcolor{purple!60}MF < HR &\cellcolor{purple!60}HF < HR \\
& VBP &- &\cellcolor{purple!60}MF > HR &\cellcolor{purple!30}HF > HR \\
& VBZ &- &\cellcolor{purple!60}MF < HR &\cellcolor{purple!15}HF < HR \\
& WDT &- &\cellcolor{purple!30}MF > HR &- \\
& WP &- &- &\cellcolor{purple!15}HF > HR \\
\hline
\multirow{7}{*}{complexity} & ttr &- &\cellcolor{purple!60}MF < HR &\cellcolor{purple!60}HF < HR \\
& avg wordlen &- &\cellcolor{purple!45}MF < HR &- \\
& word count &- &\cellcolor{purple!60}MF > HR &\cellcolor{purple!60}HF > HR \\
& flesch kincaid grade level &- &- &\cellcolor{purple!15}HF < HR \\
& smog index &\cellcolor{purple!45}HF > MF &\cellcolor{purple!60}MF < HR &- \\
& coleman liau index &\cellcolor{purple!15}HF > MF &\cellcolor{purple!60}MF < HR &- \\
& lix &- &\cellcolor{purple!45}MF < HR &\cellcolor{purple!15}HF < HR \\
\hline
\multirow{6}{*}{bias} & bias words &\cellcolor{purple!60}HF > MF &\cellcolor{purple!60}MF < HR &\cellcolor{purple!30}HF > HR \\
& factives &- &\cellcolor{purple!30}MF > HR &\cellcolor{purple!30}HF > HR \\
& implicatives &- &\cellcolor{purple!45}MF > HR &- \\
& report verbs &- &- &\cellcolor{purple!15}HF > HR \\
& positive opinion words &\cellcolor{purple!45}HF > MF &- &\cellcolor{purple!30}HF > HR \\
& negative opinion words &\cellcolor{purple!60}HF > MF &\cellcolor{purple!60}MF < HR &\cellcolor{purple!15}HF < HR \\
\hline
\multirow{8}{*}{affect} & vadneg &\cellcolor{purple!15}HF > MF &\cellcolor{purple!60}MF < HR &- \\
& vadneu &\cellcolor{purple!15}HF < MF &- &- \\
& vadpos &- &- &\cellcolor{purple!15}HF > HR \\
& wneg &\cellcolor{purple!45}HF > MF &\cellcolor{purple!45}MF < HR &- \\
& wpos &\cellcolor{purple!15}HF > MF &- &\cellcolor{purple!15}HF > HR \\
& wneu &- &- &\cellcolor{purple!45}HF > HR \\
& sneg &\cellcolor{purple!60}HF > MF &\cellcolor{purple!60}MF < HR &- \\
& sneu &- &\cellcolor{purple!30}MF > HR &\cellcolor{purple!30}HF > HR \\
\hline
\multirow{3}{*}{moral} & HarmVirtue &- &\cellcolor{purple!30}MF > HR &\cellcolor{purple!60}HF > HR \\
& AuthorityVice &- &\cellcolor{purple!15}MF < HR &- \\
& MoralityGeneral &- &\cellcolor{purple!15}MF > HR &- \\
\hline
\multirow{1}{*}{event} & num dates &\cellcolor{purple!30}HF < MF &\cellcolor{purple!60}MF > HR &- \\
\bottomrule
\end{tabular}}
\caption{Comparison of content-based features across Human-written Fake news (HF), LLM-generated Fake news (MF), and Human-written Real news (HR) for the \texttt{PolitiFact++} dataset. The table showcases differences in style, complexity, bias, affect, morale, and event features. The colour intensity represents the significance of the difference, with darker shades indicating higher significance.}
\label{tab: feature-politifact}
\end{table}

\end{document}